\documentclass[runningheads]{llncs}
\usepackage[T1]{fontenc}
\usepackage{amsmath,amssymb,amsfonts}
\usepackage{algorithm}
\usepackage{algorithmic}
\usepackage[title,toc,titletoc,header]{appendix}

\usepackage{graphicx}
\begin{document}
\title{Hierarchical Spatio-Temporal State-Space Modeling for fMRI Analysis}
%
%
\author{Yuxiang Wei\inst{1}\orcidID{0000-0001-6552-8912} \and
Anees Abrol\inst{1}\orcidID{0000-0001-9223-5314} \and
Vince D. Calhoun\inst{1}\orcidID{0000-0001-9058-0747}}
\authorrunning{Y Wei et al.}
%
\institute{Georgia Institute of Technology, Georgia State University, Emory University\\
Center for Translational Research in Neuroimaging and Data Science (TreNDS)\\
Atlanta, USA\email{\{ywei13,aabrol,vcalhoun\}@gsu.edu}}
\maketitle              
\begin{abstract}
Recent advances in deep learning structured state space models, especially the Mamba architecture, have demonstrated remarkable performance improvements while maintaining linear complexity. In this study, we introduce functional spatiotemporal Mamba (FST-Mamba), a Mamba-based model designed for discovering neurological biomarkers using functional magnetic resonance imaging (fMRI). We focus on dynamic functional network connectivity (dFNC) derived from fMRI and propose a hierarchical spatiotemporal Mamba-based network that processes spatial and temporal information separately using Mamba-based encoders. Leveraging the topological uniqueness of the FNC matrix, we introduce a component-wise varied-scale aggregation (CVA) mechanism to aggregate connectivity across individual components within brain networks, enabling the model to capture component-level and network-level information. Additionally, we propose symmetric rotary position encoding (SymRope) to encode the relative positions of each functional connection while considering the symmetric nature of the FNC matrix. Experimental results demonstrate significant improvements in the proposed FST-Mamba model on various brain-based classification and regression tasks. We further show brain connectivities and dynamics that are crucial for the prediction. Our work reveals the substantial potential of attention-free sequence modeling in brain discovery. The codes are publicly available here: \url{https://github.com/yuxiangwei0808/FunctionalMamba/tree/main}.

\end{abstract}
\section{Introduction}

The human brain is a complex organ and understanding its underlying organizations and dynamics has been an intriguing goal for neuroscientists \cite{bullmore2009complex}. Analyzing brain networks and exploiting the intrinsic connections with cognition, behaviors, and diseases is of paramount significance in biomedicine \cite{calhoun2014chronnectome,wei2023deep}. Among the popular neuroimaging modalities, functional magnetic resonance imaging (fMRI) captures the time sequence of blood oxygenation level-dependent (BOLD) signals of the whole brain and is a powerful tool for noninvasive assessment of the brain's spatiotemporal patterns. Due to the high dimensionality and noisy nature of raw fMRI data, a preprocessing pipeline is often applied to estimate the functional network connectivity (FNC) matrix either statically or dynamically by measuring the pairwise correlations between windowed time series of nodes, which are defined by a given atlas or data-dependently. Previous work has shown the association of macro-scale brain behaviors with static or dynamic FNC using statistical methods \cite{finn2015functional,mostert2016characterising,}. More recently, deep neural networks (DNN) have been widely employed to explore the nonlinear relationship between human brain cognition and disorders \cite{wei2022age,li2021braingnn,asadi2023transformer,kan2022brain,kim2023swift,kawahara2017brainnetcnn,diao2023ft2tf}.

As a powerful representation encoder, DNN is capable of disentangling latent characteristics in data and offering unique insights \cite{diao2024learning}. Popular architectures such as convolutional neural networks (CNN) and recurrent neural networks (RNN) are employed to extract spatial or temporal features. Additionally, graph neural networks (GNN) \cite{wu2020comprehensive} are proposed to analyze graph-structured data. To handle the issue of the locality of these models, a transformer is applied for fMRI exploration due to its ability to learn global representations \cite{wei2023deep}. More recently, the structured state space model (SSM), specifically Mamba \cite{gu2023mamba}, has been recognized for its ability to capture long-range dependency while maintaining linear complexity. Mamba shows superb performance in downstream tasks such as computer vision \cite{zhu2024vision}, natural language processing \cite{he2024densemamba}, and graph learning \cite{wang2024graph}.

Brain functional network connectivity has unique characteristics that make Mamba-based models ineffective. The FNC matrix is an $N \times N$ symmetric matrix, where each row or column represents a brain component's correlations with all $N$ components. Adjacent components can be grouped into functional networks due to shared activation and deactivation patterns, making both component-level and network-level topological priors important. Treating the matrix as an image or flattening it loses intrinsic topological relationships \cite{caramazza2006cognitive}. Additionally, dynamic FNC (dFNC) captures crucial cognitive and behavioral information \cite{allen2014tracking}. While spatiotemporal Mamba extensions exist \cite{li2024videomamba,yang2024vivim}, they work with clearer contextual information and higher SNR than noisy fMRI signals, which may include artifact-generated dFNC frames \cite{liu2016noise}. Finally, Mamba's sensitivity to token scan order does not reflect actual functional relationships and standard scanning strategies can obscure patterns due to the FNC matrix's symmetry. Thus, additional modeling strategies are needed to encode this information effectively.

In this work, we propose \textbf{F}unctional \textbf{S}patio-\textbf{T}emporal \textbf{Mamba} (FST-Mamba), which leverages the unique properties of FNC to fully exploit the power of Mamba in brain network analysis. Specifically, we adopt a hierarchical structure to capture multi-scale space and time information. Accounting for the topological priors from functional components and networks, we propose the component-wise varied-scale aggregation (CVA) mechanism to aggregate component-wise connections from networks across the brain and enhance Mamba's ability to capture global interdependencies.  Additionally, to maximally extract intra-network information while reducing interference, we propose the component-specific selective scan to model per-component information. Finally, we propose symmetrical rotary position encoding (SymRope) to encode relative positional relationships between connections, while tailoring for the symmetric property of FNC.

We present extensive experiments to verify our designs based on three benchmark datasets: the Human Connectome Project (HCP) \cite{van2013wu}, the UK BioBank (UKB) \cite{miller2016multimodal}, and Alzheimer's Disease Neuroimaging Initiative \cite{jack2008alzheimer}. We conduct experiments based on several classification and regression tasks, including sex, dementia, age, and fluid intelligence classification. The results show that our FST-Mamba significantly outperforms other baseline methods. Furthermore, we interpret the models by showing the key connectivities and dynamics that contribute to the prediction, suggesting that the model can uncover fMRI-based biomarkers for demographics and cognitive diseases.

\section{Related Work}

\subsection{Dynamic Brain Network Analysis}
Previous works generally fall into two research lines to analyze and extract useful information from dynamic brain networks: graph-based  and imaging-based. For the former, Li et al. \cite{li2021braingnn} designed a component-aware GNN to learn the functional connections and proposed a pooling operator to select key components. Kan et al. \cite{kan2022brain} adapted the graph transformer and proposed a readout to gather information from the nodes. However, graph-based methods usually require additional pre-processing to ensure the data is in the correct form. For the latter, Asadi et al. \cite{asadi2023transformer} utilized the transformer to learn spatiotemporal representations from fMRI. Kim et al. \cite{kim2021representation} employed the variational auto-encoder to learn and generate representational geometry from fMRI. The proposed FST-Mamba falls in between two research lines, where we utilize the novel state-space model and consider the unique topological properties of FNC.

\subsection{State Space Models for Brain Discovery}
The state space model (SSM) is a linear time-invariant sequence model that maps a sequence $x_t\in\mathbb{R}^L$ to its response $y_t \in\mathbb{R}^L$ with a corresponding latent space $h_t\in\mathbb{S}^{N\times L}$. Specifically, SSM can be described as a linear ordinary differential equation with an evolution parameter $A \in \mathbb{R}^{N \times N}$, projection parameters $B, C \in \mathbb{R}^N$ for a state size $N$, and the skip connection $D \in \mathbb{R}^1$. Recent work  \cite{gu2020hippo} proposes to discretize the system with a time scale parameter $\Delta$:
\begin{equation}
\begin{aligned}
    &h_t = \bar{A}h_{t-1} + \bar{B}x_t \\
    &y_t = Ch_t + Dx_t
\end{aligned}
\label{eqSSMDisc}
\end{equation}
where
\begin{equation}
\begin{aligned}
    &\bar{A} = {\rm exp}(\Delta A) \\
    &\bar{B} = (\Delta A)^{-1}({\rm exp}(\Delta A) - \bold{I})\cdot\Delta B
\end{aligned}
\end{equation}

Although the vanilla SSM Eq. \eqref{eqSSMDisc} achieves linear complexity and can be scaled up efficiently, the lack of contextual awareness hinders its performance on discrete and information-dense data \cite{gu2023mamba}. To resolve this, \cite{gu2023mamba} proposed the Mamba model that improves the SSM by introducing a data-dependent selective recurrent scan mechanism, which makes $\Delta$, $B$, and $C$ parameters trainable to filter out irrelevant information.

Mamba has gained much popularity in medical diagnosis and brain discovery. Ji and colleagues \cite{ji2024deform} proposed a Mamba-based encoder-decoder structure to generate high-quality MRI images. A recent work \cite{wang2024mamba} adapted vision Mamba in the U-Net fashion and designed a new model for medical image segmentation. While most previous works in the field of biomedicine explore Mamba for 2D imaging, few of them consider Mamba's ability to model temporal variations of highly structured data.  In this work, we study Mamba for the dynamic functional network connectivity (dFNC) matrix calculated from fMRI, where each entry represents the dynamic correlation between two function components. 

\begin{figure*}[h]
\centering
\includegraphics[width=\textwidth]{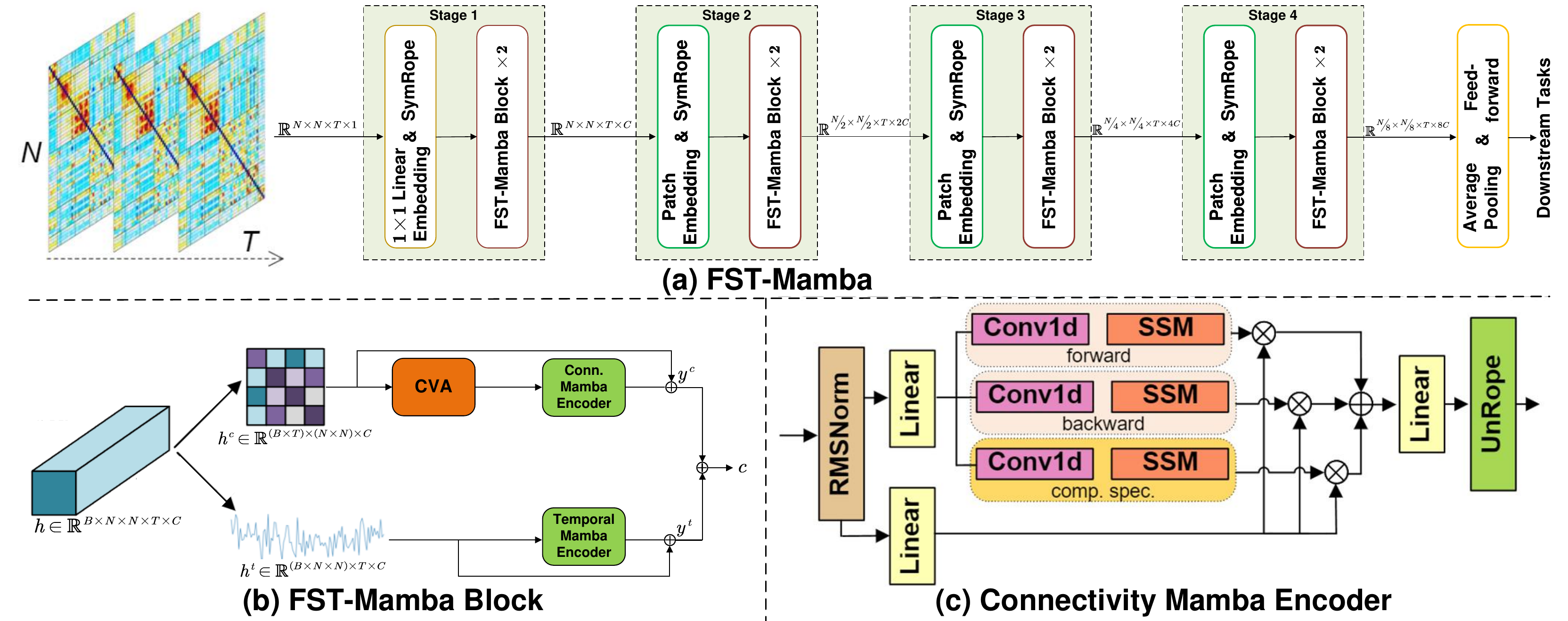}
\caption{(a) The overall architecture of the proposed FST-Mamba. (b) The fundamental building block of FST-Mamba. Instead of processing space and time simultaneously, FST-Mamba applied separate branches to reduce computational costs while effectively extracting the dynamics of fMRI. (c) FST-Mamba processes the sequence in both forward and backward directions. Additionally, the connectivity mamba encoder further processes the projected sequence in a component-specific order.}
\label{fig:framework}
\end{figure*}

\section{Methodology}
In this section, we delineate the core modules and designs of FST-Mamba for brain discovery. The overall architecture of the proposed method is depicted in Fig. \ref{fig:framework}. FST-Mamba consists of 4 stages that gradually extract relevant information from coarse to fine. The first stage consists of a linear embedding (which is a $1\times1$ convolution that expands the channel dimension to $C$), followed by 2 consecutive FST-Mamba blocks. Then, a patch merging layer is applied for the next 3 stages for downampling. Note that for each stage, we apply the symmetric rotary position encoding (SymRope) to encode valuable topological information and perform "unRope" after each stage to prevent artifacts. Finally, we perform average pooling over the feature map from the last stage for downstream tasks.

\subsection{FST-Mamba Block}

We propose to individually extract temporal and spatial information to avoid information loss. As in Fig. \ref{fig:framework}(b), assume input $h \in \mathbf{R}^{B \times N \times N \times T \times C}$, where $B$ is the batch size, $N$ is the component amount, $T$ is the temporal dimension, and $C$ is the channel amount. We merge the space or time dimension into the batch for the corresponding branch, thereby allowing Mamba to process 4D spatiotemporal input. Note that the varied-scale aggregation (CVA) module is employed to aggregate components' features across networks in the spatial branch. The procedure can be described as:

\begin{equation}
\begin{aligned}
    &y^c = \mathrm{ConnMamba}(\mathrm{CVA}(h^c)) + h^c, \\
    &y^t = \mathrm{TempMamba}(h^t) + h^t \\
    &z = y^c + y^t
\end{aligned}
\end{equation}

\subsubsection{Component-Wise Varied-Scale Aggregation}
Assume the input size of $x \in \mathbb{R}^{B\times N \times N}$, CVA transforms it into $\mathbb{R}^{(B \times k) \times \frac{N}{k} \times N}$. Specifically, it applies a sliding window $w \in \mathbb{R}^{k \times 1}$ of step size $k$ and produces $N \parallel k $ windows. The $i$-th element of each group is grouped, creating $k$ sequences where each sequence contains connectivity features across all networks. By reducing the relative distance during sequential scanning, CVA allows the Mamba encoder to effectively capture inter-network connectivity correlations. Note that we adapt the step size of CVA by halving it after each stage to be in accord with the dimension.

\begin{figure}[h]
\centerline{\includegraphics[width=\columnwidth]{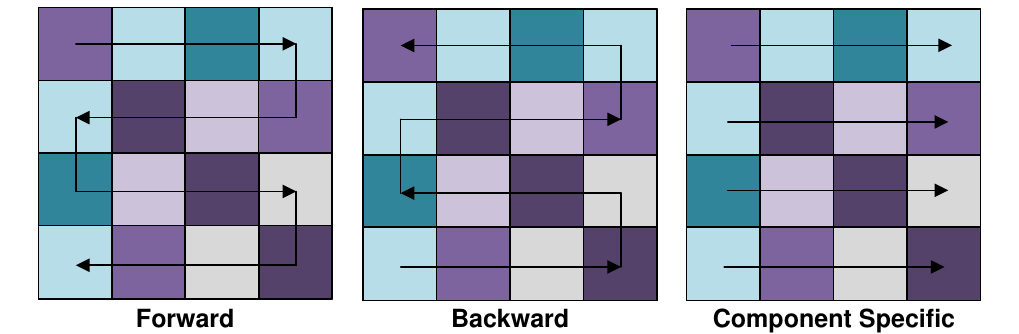}}
\caption{The forward, backward, and the proposed component-specific scan order}
\label{fig:scan}
\end{figure}

\subsection{Connectivity and Temporal Mamba Encoder}
We use a common backbone for the temporal and connectivity Mamba encoder. As illustrated in Fig. \ref{fig:framework}(c), following the previous work \cite{zhu2024vision}, we apply SSM in both the forward and backward directions to capture long-range dependencies in space and time. This allows the model to obtain bidirectional dependencies across space and time. Moreover, since an individual component connectivity profile can be significant to the brain function,  we propose a component-specific selective scan to maximally preserve information from each component. The illustration is shown in \ref{fig:scan}. Instead of directly flattening the matrix and performing scanning, we only scan the connectivity within a component, such that each row in the FNC matrix is regarded as a complete sequence. Note that the component-specific SSM is only applied for the connectivity Mamba encoder.

\subsection{Symmetric Rotary Position Encoding}
Unlike CNN which is translational-invariant, Mamba performs sequence modeling to patches and therefore lacks positional sensitivity. This necessitates an additional embedding to encode the location-related information.

Rotary position encoding (Rope) \cite{su2024roformer} encodes the absolute positions with a rotary matrix and endows relative positional dependencies to the original representation. Assume two arbitrary vectors $ \boldsymbol{m}=\mathbb{R}^m$ and $\boldsymbol{n}=\mathbb{R}^n$, Rope applies a rotary matrix $\boldsymbol{\mathcal{R}}$ such that:
\begin{equation}
   (\boldsymbol{\mathcal{R}}_m \boldsymbol{m})^{\top}(\boldsymbol{\mathcal{R}}_n \boldsymbol{n}) =  \boldsymbol{n}^{\top} \boldsymbol{\mathcal{R}}_m^{\top}\boldsymbol{\mathcal{R}}_n \boldsymbol{n} = \boldsymbol{m}^{\top} \boldsymbol{\mathcal{R}}_{n-m} \boldsymbol{n}\
\end{equation}
where

\begin{equation}
\begin{aligned}
    &\boldsymbol{\mathcal{R}}_n=\begin{pmatrix}\cos n\theta & -\sin n\theta\\ \sin n\theta & \cos n\theta\end{pmatrix}=\exp\left\{n\boldsymbol{B}\right\},\\
    &\boldsymbol{B}=\begin{pmatrix}0 & -\theta\\ \theta & 0\end{pmatrix}
\end{aligned}
\label{eq:1drope}
\end{equation}

To extend to the 2D case, an intuitive solution for the $\boldsymbol{\mathcal{R}}$ can be:
\begin{equation}
    \boldsymbol{\mathcal{R}}_{x,y}=\exp \big(x\boldsymbol{B}_1 + y\boldsymbol{B}_2\big)
\end{equation}

To fulfill the property of $\boldsymbol{\mathcal{R}}$ and encode the relative positions, we need $\boldsymbol{\mathcal{R}}_{x_1,y_1}^{\top}\boldsymbol{\mathcal{R}}_{x_2,y_2}=\boldsymbol{\mathcal{R}}_{x_2-x_1,y_2-y_1}$, for which previous work \cite{su2024roformer} offers a straightforward solution:

\begin{equation}
\begin{aligned}
    &\boldsymbol{\mathcal{R}}_{x,y} = \exp \left\{\begin{pmatrix}
    0 & -x & 0 & 0 \\
    x & 0 & 0 & 0 \\
    0 & 0 & 0 & -y \\
    0 & 0 & y & 0
    \end{pmatrix} \theta \right\} \\
    &= \begin{pmatrix}
    \cos x \theta & -\sin x \theta & 0 & 0 \\
    \sin x \theta & \cos x \theta & 0 & 0 \\
    0 & 0 & \cos y \theta & -\sin y \theta \\
    0 & 0 & \sin y \theta & \cos y \theta
    \end{pmatrix}
\end{aligned}
\label{eq:ropeR}
\end{equation}

However, since the FNC matrix is symmetric along its diagonal, symmetrical entries should be encoded with the same positional information, i.e., $\boldsymbol{\mathcal{R}}_{x,y}=\boldsymbol{\mathcal{R}}_{y,x}$. This necessitates the additional constraint:

\begin{equation}
    \boldsymbol{\mathcal{R}}_{x,y}\boldsymbol{\mathcal{R}}^{\top}_{x,y} = \boldsymbol{\mathcal{R}}^{\top}_{x,y}\boldsymbol{\mathcal{R}}_{x,y} = I
\end{equation}

To accommodate this constraint, we adapt Eq. \ref{eq:ropeR} to:

\begin{equation}
    \hat{\boldsymbol{\mathcal{R}}}_{x,y} = \left(\begin{pmatrix}
    \sin x\theta & \cos x\theta & 0 & 0 \\ 
    \cos x\theta & -\sin x\theta & 0 & 0 \\ 
    0 & 0 & \sin y\theta & \cos y\theta \\ 
    0 & 0 & \cos y\theta & -\sin y\theta \\ 
\end{pmatrix}\right)
\label{eq:symRx}\end{equation}

At each FST-Mamba stage, we separately apply Eq. \ref{eq:ropeR} on the temporal dimension and Eq. \ref{eq:symRx} on the spatial dimension, thereby independently encoding positional information in space and time. Furthermore, positional encoding is performed at each stage after downsampling, which can introduce artifacts to the representation. Therefore, we propose to "unRope" the representations at the output of each stage by applying the inverse of $\hat{\boldsymbol{\mathcal{R}}}$ and $\boldsymbol{\mathcal{R}}$. Note that the $\hat{\boldsymbol{\mathcal{R}}}^{-1} = \hat{\boldsymbol{\mathcal{R}}}$ and:
\begin{equation}
    \hat{\boldsymbol{\mathcal{R}}}_{x,y}^{-1} = \left(\begin{pmatrix}
    \cos x\theta & \sin x\theta & 0 & 0 \\ 
    -\sin x\theta & \cos x\theta & 0 & 0 \\ 
    0 & 0 & \cos y\theta & \sin y\theta \\ 
    0 & 0 & -\sin y\theta & \cos y\theta \\ 
\end{pmatrix}\right)
\label{eq:Rxinv}\end{equation}

\section{Experiments}

\subsection{Experimental Settings}

\subsubsection{Datasets and Preprocessing}

We use the dFNC data of 833 healthy young subjects from the Human Connectome Project (HCP) S1200 data \cite{van2013wu} to predict sex and age,  12626 adults from the UK BioBank (UKB) \cite{sudlow2015uk} to predict sex and fluid intelligence and 474 subjects from the ADNI dataset \cite{jack2008alzheimer} to diagnose dementia. We use accuracy (ACC) and area under ROC curve (AUC) as the metrics for classification and mean absolute error (MAE) and mean squared error (MSE) as the metrics for regression.

After preprocessing the fMRI data, we employ the spatially constrained independent component analysis (ICA) \cite{calhoun2001method} with the Neuromark\_fMRI\_1.0 template \cite{du2020neuromark}, which generates 53 brain components divided into 7 networks. To compute the dynamic functional network connectivity (dFNC), we apply the sliding window approach and compute the pair-wise Pearson correlation between 53 components in each window.

\subsubsection{Baselines}
We use several recent deep learning networks that are proposed for fMRI-based brain network analysis, including static brain network methods: Brain Network Transformer (BNT) \cite{kan2022brain}, BrainNetCNN \cite{kawahara2017brainnetcnn}, and FBNETGEN \cite{kan2022fbnetgen}. We further include dynamic brain network methods: SwiFT \cite{kim2023swift}, CNNLSTM \cite{qayyum2022efficient}, and FE-STGNN \cite{chen2023fe}. Additionally, we also include the vanilla Mamba \cite{gu2023mamba} that directly takes the flattened dFNC. Detailed descriptions of the baselines are provided in the appendix.

\subsubsection{Implementation Details}
For FST-Mamba, we use the same architecture across all the experiments, with the channel number $C=24$. For training, we employ AdamW \cite{loshchilov2017decoupled} with a batch size of 64 and a learning rate of 0.001. We train the proposed method and other baselines for 200 epochs using cosine annealing.

\begin{table*}[h]
\caption{Performance comparison on classification and regression tasks. We employ paired t-test to test significance, where $^{**}$ represents $p < 0.01$ and $^{*}$ represents $p < 0.05$}
\centering
\begin{tabular}{c|cc|cc|cc|cc|cc}
\hline
                   & \multicolumn{2}{c|}{HCP-sex}  & \multicolumn{2}{c|}{HCP-age}  & \multicolumn{2}{c|}{UKB-sex}  & \multicolumn{2}{c|}{UKB-fluid} & \multicolumn{2}{c}{ADNI}      \\ \hline
                   & ACC           & AUC           & MAE           & MSE           & ACC           & AUC           & MAE            & MSE           & ACC           & AUC           \\ \hline
BNT                & 78.9$^{**}$          & 79.6$^{**}$          & 2.93          & 13.0$^{*}$          & 72.7$^{**}$          & 82.3$^{**}$          & 1.58$^{*}$           & 4.12$^{**}$          & 85.7$^{**}$          & 79.8$^{*}$          \\
BrainNetCNN        & 82.0$^{*}$          & 86.9          & 3.90$^{**}$          & 18.2$^{**}$          & 84.9          & 91.8$^{*}$          & 1.72$^{**}$           & 6.31$^{**}$          & 85.9$^{**}$          & 77.8$^{**}$          \\
FBNETGEN           & 80.5$^{**}$          & 86.4$^{*}$          & 2.95$^{*}$          & 12.8$^{*}$          & 84.9          & 91.5$^{*}$          & 1.59$^{**}$           & 3.99$^{**}$          & 86.1$^{*}$          & 79.6$^{*}$          \\
SWiFT              & 72.9$^{**}$          & 71.6$^{**}$          & 2.95$^{*}$          & 13.2$^{**}$          & 83.5$^{*}$          & 91.4$^{*}$          & 1.60$^{**}$           & 4.16$^{**}$          & 71.9$^{**}$          & 65.7$^{**}$          \\
CNNLSTM            & 74.5$^{**}$          & 74.4$^{**}$          & 3.09$^{**}$          & 13.5$^{**}$          & 84.4$^{*}$          & 91.7$^{*}$          & 1.61$^{**}$           & 4.12$^{**}$          & 84.0$^{**}$          & 68.2$^{**}$          \\
FE-STGNN           & 71.9$^{**}$          & 74.0$^{**}$          & 3.19$^{**}$          & 14.5$^{**}$          & 76.7$^{**}$          & 84.5$^{**}$          & 1.57$^{*}$           & 3.91$^{**}$          & 83.3$^{**}$          & 64.1$^{**}$          \\
Mamba              & 69.8$^{**}$& 73.6$^{**}$          & 3.03$^{**}$          & 13.2$^{**}$          & 82.4$^{**}$          & 89.8$^{**}$          & 1.61$^{**}$           & 4.20$^{**}$          & 83.6$^{**}$          & 70.9$^{**}$          \\ \hline
\textbf{FST-Mamba} & \textbf{83.5}& \textbf{87.5} & \textbf{2.90}& \textbf{12.1}& \textbf{85.3} & \textbf{92.7} & \textbf{1.50}  & \textbf{3.82} & \textbf{87.1} & \textbf{81.9}
\end{tabular}\label{tab:perform}\end{table*}

\subsection{Performance Comparison}
We compare the proposed FST-Mamba with baselines on various classification and regression tasks. As in Table \ref{tab:perform}, on the sex and dementia classification tasks, the proposed FST-Mamba outperforms all baselines on HCP, UKB, and ADNI datasets. On the regression tasks, FST-Mamba shows competitive results against the best-performing baseline model, such as FBNETGEN. Meanwhile, the proposed model exhibits notable enhancement in sex and dementia classification. It should be noted that some baseline models show distinct performance on different tasks. For instance, BrainNetCNN is the best baseline model for HCP sex prediction, whereas it is the worst for HCP age prediction. Furthermore, baselines using dynamic brain networks are not necessarily better than static methods, possibly due to the noisy nature of fMRI signals. Finally, the standard Mamba, while achieving reasonable results on the ADNI and HCP datasets, performs poorly on the HCP dataset.

\subsection{Ablation Studies}
We conduct extensive ablation studies to validate the designs of the proposed model based on the HCP dataset. The results are presented in Table \ref{tab:ablMamba} and \ref{tab:ablPos}. From the table, we observe that both the connectivity Mamba and the temporal Mamba branches distill highly predictive signals. As seen in Table \ref{tab:ablMamba}, removing the connectivity Mamba significantly degrades the performance, whereas the temporal Mamba is comparatively lesser predictive.

\begin{table}[t]
\caption{Ablation study for connectivity and temporal mamba}
\centering
\begin{tabular}{c|cc|cc}
                   & \multicolumn{2}{c|}{HCP-sex}  & \multicolumn{2}{c}{HCP-age}  \\ \hline
                   & ACC           & AUC           & MAE           & MSE           \\ \hline
w/o Conn. Mamba    & 70.5          & 68.7          & 3.06          & 13.4          \\
w/o Temporal Mamba & 81.1          & 85.9          & 2.98          & 12.8          \\ \hline
w/o CVA             & 81.4          & 86.1          & 2.95          & 12.8          \\
w/o Comp. Spec. scan & 80.8          & 86.2          & 2.97          & 12.6          \\ \hline
\textbf{FST-Mamba}  & \textbf{83.5} & \textbf{87.5} & \textbf{2.90}& \textbf{12.1}\end{tabular}
\label{tab:ablMamba}
\end{table}

\begin{table}[h]
\caption{Ablation study for the positional encoding}
\centering
\begin{tabular}{c|cc|cc}
                 & \multicolumn{2}{c|}{HCP-sex}  & \multicolumn{2}{c}{HCP-age}   \\ \hline
                 & ACC           & AUC           & MAE           & MSE           \\ \hline
w/o Pos. Enc.    & 77.8          & 76.1          & 3.06          & 13.2          \\
Abs. Pos.        & 80.6          & 82.7          & 2.98          & 12.9          \\
w/o unRope       & 82.5          & 87.0          & 2.95          & 12.7          \\ \hline
\textbf{SymRope} & \textbf{83.5} & \textbf{87.5} & \textbf{2.90}& \textbf{12.1}\end{tabular}
\label{tab:ablPos}
\end{table}

We further show the necessity of positional encoding, the proposed SymRope (substitute SymRope with absolution positional encoding \cite{vaswani2017attention}), and "unRope" after encoding. The results are presented in Table \ref{tab:ablPos}. Positional encoding can be significant to FST-Mamba since vnilla Mamba lacks positional awareness.

\begin{figure*}[h]
\centerline{\includegraphics[width=\textwidth]{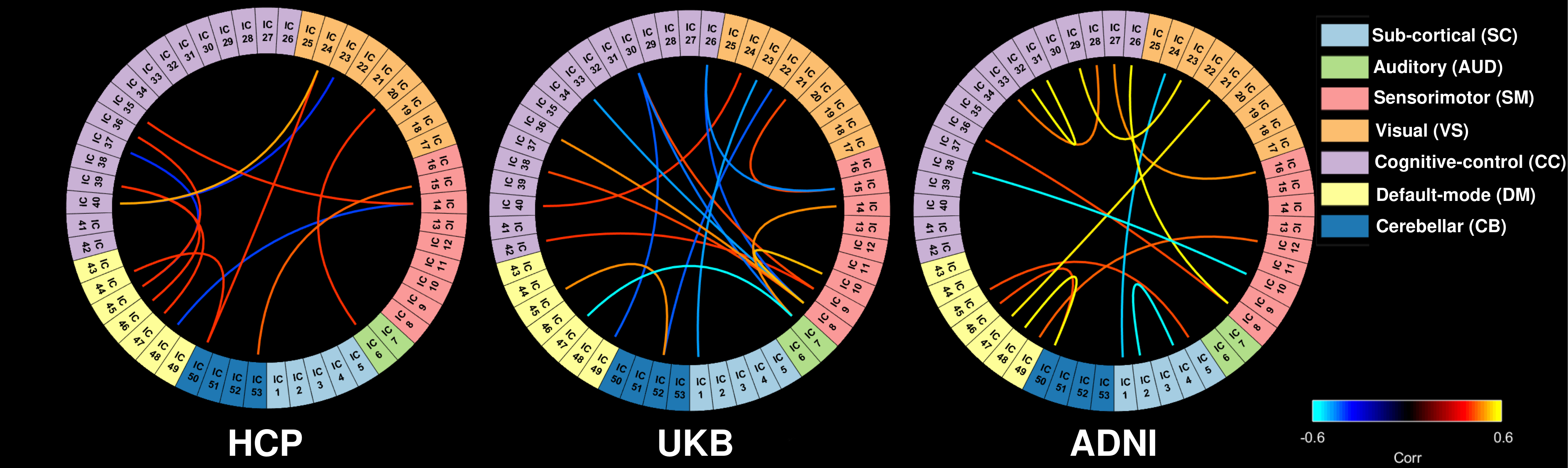}}
\caption{Crucial connections for sex classification (HCP and UKB datasets) and dementia classification (ADNI dataset)}
\label{fig:viz}
\end{figure*}

\begin{figure*}[h]
\centerline{\includegraphics[width=\textwidth]{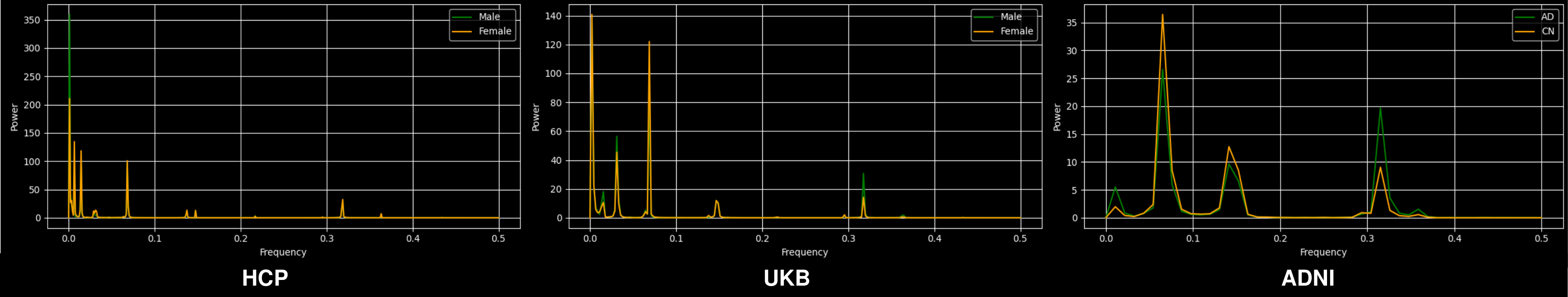}}
\caption{Periodograms for sex classification (Male vs. Female) and dementia classification (CN vs. AD). Note that the frequency is normalized.}
\label{fig:vizT}
\end{figure*}

\subsection{Interpretation Results}
Using the GradientShap \cite{lundberg2017unified} method implemented in Captum \cite{kokhlikyan2020captum}, we identify brain connections and dynamics that show significant explanatory power in dementia and sex classification tasks. For the spatial analysis, we calculate the average difference between the positive and negative classes' connectograms and apply a threshold of 0.4 to retain only the crucial connections. The results are shown in Fig. \ref{fig:viz}. When predicting sex, the model emphasizes inter-network connections involving CC and SM. Notably, AUD and DM exhibit additional negative correlations with predictions on the UKB dataset. In contrast, for the ADNI dataset, the model focuses on intra-network connections, primarily within DM and CC networks, while also showing negative correlations with CC-SM and SC-VS connections. Interestingly, connections such as CC-SM contribute to both sex and dementia predictions, suggesting an underlying relationship between sex and dementia \cite{ruitenberg2001incidence}.

To show the dynamics learned by the model, we present key frequency bands by plotting the periodograms for each dataset. As in Fig. \ref{fig:vizT}, the model is particularly attentive to the low-frequency band ($f < 0.2$) on sex classification, where several spikes display strong power. Conversely, in the ADNI dataset, the power is more evenly distributed across high- and low-frequency components, both of which are crucial for prediction. Additionally, there is a notable difference in power between positive and negat ive classes in the high-frequency band ($f > 0.3$) on both UKB and ADNI datasets. This suggests a correlation between sex and dementia concerning the brain's temporal activities.

\section{Conclusion}

Analyzing the spatiotemporal dynamics of functional networks using dynamic functional network connectivity is challenging due to its topological uniqueness and dimensionality. In this paper, we present FST-Mamba, a Mamba-based spatiotemporal network designed for high-dimensional dFNC features that can learn dynamic connectivity patterns and predict biological and cognitive outcomes. We present extensive experiments to verify the effectiveness of our models on multiple tasks and uncover crucial brain connections and dynamics that contribute to the classification. For future work, we plan to explore the scalability of FST-Mamba and improve it with pretraining.

%
%
%
\bibliographystyle{splncs04}
\bibliography{ref}

\end{document}